\documentclass[runningheads]{llncs}

\usepackage{eccv}

\usepackage{eccvabbrv}

\usepackage{graphicx}
\usepackage{booktabs}

\usepackage{amsmath}
\usepackage{amssymb}
\usepackage{bm}
\usepackage{xcolor}
\usepackage{tikz}

\usetikzlibrary{
    arrows.meta,
    positioning,
    calc,
    fit,
    backgrounds
}

\usepackage[accsupp]{axessibility}  

\usepackage{xcolor}
\usepackage{tikz}
\usetikzlibrary{arrows.meta,positioning,calc,fit,backgrounds}

\usepackage{hyperref}

\usepackage{orcidlink}

\newcommand{\cofirst}{\textsuperscript{*}}

\begin{document}

\title{FaLCon: Facet-Anchored Retrieval with Late Consensus for Sim2Real Text-Based Person Anomaly Search} 

\titlerunning{FaLCon}

\author{
Hieu Dinh Trung Pham\inst{1}\cofirst
\and
Phuong Huu Vu Tran\inst{1}\cofirst \\
Thuan Duc Mai\inst{2} \and Son Nguyen Minh Le\inst{1} \and Khang Le Minh\inst{1} \\
\and Hoang Vo \inst{4,5}  \and Minh-Chi Phung \inst{3,5} \\
Huy Minh Nhat Nguyen\inst{1} \and Cuong Tuan Nguyen\inst{1}
}

\authorrunning{H.~D.~T.~Pham et al.}

\institute{ PAMI Lab, Vietnamese-German University, Vietnam \and
Ho Chi Minh City University of Technology, Vietnam \and
University of Information Technology, Ho Chi Minh City, Vietnam \and
Ho Chi Minh city University of Science, Vietnam \and
GenAI4E Lab
}
\maketitle

\begingroup
\renewcommand{\thefootnote}{*}
\footnotetext{Hieu Dinh Trung Pham and Phuong Huu Vu Tran contributed equally to this work.}
\endgroup

\begin{abstract}
Text-based person anomaly search requires retrieving real-world pedestrian images from detailed natural-language descriptions using models trained primarily on synthetic data. This Sim2Real setting is particularly challenging because visually similar candidates may differ only in subtle actions, object interactions, or appearance attributes, while applying multimodal large language models to the entire gallery is computationally expensive. We propose an anchor-constrained coarse-to-fine retrieval framework that combines global semantic matching with fine-grained verification. First, each query is represented by its original caption, a structured concatenation, and several semantic facets. Heterogeneous vision-language retrievers are then integrated through robust per-query score calibration and soft claim-aware fusion. Full and concatenated captions serve as anchors to preserve candidate recall, whereas appearance, action, and object facets provide bounded corrective evidence. The resulting candidate pool is further refined by a discriminative Qwen3 reranker and two complementary semantic verification modules based on anomaly-aware cloze completion and multi-agent evidence reasoning. Finally, an uncertainty-gated consensus module adaptively reweights the three experts on ambiguous queries. Experiments on the PAB benchmark show that the proposed soft claim-aware retrieval achieves 86.44\% mAP@10, substantially outperforming individual retrieval backbones. The complete framework further improves performance to 95.41\% mAP@10, 94.44\% R@1, and 99.09\% R@5. These results demonstrate that preserving strong global retrieval while restricting expensive semantic reasoning to a small candidate pool is effective for fine-grained Sim2Real person anomaly search. Our code will be available on \href{https://github.com/hieupham1103/AI-CITY-10th-ECCV2026}{Github}.
\end{abstract}
\section{Introduction}

Text-based person retrieval aims to identify a target pedestrian from a large image gallery using a natural-language description, which is particularly useful when no visual query is available~\cite{li2017person}. Although recent vision--language models have substantially improved appearance-based matching~\cite{radford2021learning,jiang2023irra}, existing benchmarks mainly emphasize clothing and routine actions. Consequently, models often capture what a person looks like more reliably than what the person is doing.

The Pedestrian Anomaly Behavior (PAB) benchmark extends this setting to \emph{text-based person anomaly search}, where queries describe appearance, actions, object interactions, and surrounding scenes~\cite{yang2024beyond}. AI City Challenge 2026 Track~4 adopts more than one million synthetic image-text pairs for training and evaluates on real-world images, creating a challenging Sim2Real setting~\cite{aicity2026track4}. The test gallery contains $1{,}978$ target images and $34{,}795$ distractors, many of which share similar appearance, pose, or scene context.

This task presents two major challenges. First, a global image-text embedding may overlook subtle distinctions in actions, attributes, or object interactions. Second, directly applying multimodal large language models to the complete gallery is computationally impractical. These observations motivate a coarse-to-fine design that preserves high candidate recall during retrieval and restricts expensive semantic reasoning to a small set of difficult candidates~\cite{ju2025anomalylmm,xie2026ssdc}.

We propose a multi-expert coarse-to-fine framework. Each query is represented using its full caption, a structured concatenation, and several semantic facets. Global caption branches serve as anchors, while appearance, action, and object branches provide bounded corrective evidence through robust score calibration and soft claim-aware fusion. The resulting candidate pool is then processed by three complementary reranking experts: a discriminative Qwen3 reranker, anomaly-aware cloze completion, and multi-agent semantic verification. Finally, an uncertainty-gated consensus rule adaptively combines their predictions when the fused prediction is uncertain. Unlike generative debate methods~\cite{verma2026selene,eo2025down,fan2026imad}, our experts do not exchange textual arguments; the consensus module only reweights scores that have already been computed.

Our main contributions are:
\begin{itemize}
    \item We propose anchor-constrained soft claim-aware retrieval that combines global semantics with fine-grained appearance, action, and object cues while preventing unsupported facet-only candidates.

    \item We build a coarse-to-fine cascade that applies discriminative and generative semantic verification only to a compact candidate pool.

    \item We introduce uncertainty-gated consensus to resolve expert conflicts without extra multimodal inference, achieving $95.4078\%$ mAP@10, $94.4388\%$ R@1, and $99.0900\%$ R@5 on PAB.
\end{itemize}

\section{Related Work}
\subsection{Text-Based Person Retrieval and Anomaly Search}
\label{sec:related_tbpr}
 
\noindent\textbf{Text-based person re-identification.}
Text-based person re-identification retrieves the pedestrian image matching a
natural-language query, a task established by CUHK-PEDES~\cite{li2017person}.
Early methods align separately trained encoders via explicit part/attribute
matching~\cite{wang2020vitaa,chen2022tipcb,zhang2018deep}, but incur a modality
gap and rely on external priors that distort intra-modality
cues~\cite{jiang2023irra}. CLIP-based frameworks~\cite{radford2021learning}
instead learn implicit cross-modal relations on a jointly pre-trained
backbone~\cite{jiang2023irra}.
 
\noindent\textbf{PAB and pose-aware retrieval.}
Such methods align \emph{static} appearance and overlook motion.
PAB~\cite{yang2024beyond} reframes the task as anomaly search, describing
appearance \emph{and} behavior (e.g., falling, being hit), with $\sim$1M
synthetic training pairs and $1{,}978$ real test pairs, and its CMP baseline
injects pose~\cite{cao2017realtime,jing2020pose} with hard negatives to separate
normal from anomalous actions. Yet these cues are transient and fine-grained,
and pose signals degrade under the domain shift the benchmark introduces.
 
\noindent\textbf{Prior challenge solutions.}
On the WWW\,2025 challenge, top entries continuously pre-train
EVA-CLIP/OpenCLIP~\cite{sun2023evaclip,ilharco2021openclip} with ensembling, or
fine-tune multi-grained X-VLM~\cite{zeng2022xvlm} with hard-negative
sampling~\cite{he2025efficient,bai2023rasa}. These gains, however, lean on heavy
pre-training and ensembles, while local--global and part-slot
modeling~\cite{niu2020improving,park2024plot} sharpen fine-grained alignment only
at added matching cost.
 
\noindent\textbf{Sim2Real and anomaly behavior.}
Since PAB trains on synthetic data yet tests on real footage, un-adapted
backbones collapse (EVA-CLIP at $60.0$ Recall@1 zero-shot)~\cite{he2025efficient},
and long-tailed anomalies make supervised fine-tuning over-fit ``normal''
behavior~\cite{ju2025anomalylmm}. Recent responses target these axes via
training-free LMM adaptation~\cite{ju2025anomalylmm}, synthetic-data
curation~\cite{sun2025filtering}, and links to classic video anomaly
detection~\cite{sultani2018real}, framing the fine-grained, Sim2Real-robust, long-tail-reliable retrieval this work targets.

\subsection{Fine-Grained Vision-Language Retrieval}
\label{sec:related_finegrained}

Dual-encoder vision-language models enable efficient large-scale retrieval,
but their global embeddings may overlook subtle differences in attributes,
object interactions, and human actions. This limitation is critical for person
anomaly search, where candidates often share similar appearance and scene
context but differ in a small behavioral cue. Prior work improves
discrimination through local--global visual representations, cross-modal
interaction objectives, and hard-negative learning
\cite{nguyen2025hybrid,yang2024beyond,hu2025solution}. In particular,
VisMin shows that vision-language models remain vulnerable to minimal changes
in objects, attributes, counts, and spatial relations, while training with such
hard negatives improves fine-grained alignment~\cite{vismin}.

These findings motivate combining complementary representations rather than
relying on a single global similarity. Our method therefore retrieves
candidates using both full-caption and facet-specific embeddings, with
fine-grained action, appearance, and object signals acting as bounded
corrections to the stronger global branches.

\subsection{Coarse-to-Fine Multimodal Reranking}
\label{sec:related_reranking}
Coarse-to-fine retrieval first uses an efficient embedding model to reduce the
gallery to a compact candidate set, and then applies a more expressive
cross-modal model to resolve difficult matches. This design preserves the
scalability of dual-encoder retrieval while allowing deeper image--text
interaction only for top-ranked candidates. AnomalyLMM follows this paradigm
by masking verbs and color attributes in the query, completing them from visual
evidence, and semantically re-ranking a small candidate list
\cite{ju2025anomalylmm}. SSDC further decouples structural retrieval from
semantic verification, using a Detective--Analyst--Writer workflow to filter
negatives, extract evidence, and synthesize candidate-specific descriptions
\cite{xie2026ssdc}.

\noindent\textbf{Selective multi-agent debate.}
Multi-agent debate reconciles disagreement between models by letting them
critique one another~\cite{du2024improving,liang2024encouraging}, and recent
work invokes it only when needed: SELENE gates on semantic disagreement and
confidence--likelihood misalignment~\cite{verma2026selene}, DOWN on the initial
confidence score~\cite{eo2025down}, and iMAD on learned hesitation
features~\cite{fan2026imad}. All three gate primarily to reduce generation cost; iMAD additionally gates for accuracy, and all
operate on generative agents.

Unlike approaches that apply a single semantic verifier after coarse retrieval,
our framework evaluates the same top-$10$ candidate pool with three parallel
experts: a PAB-adapted Qwen3 reranker, an AnomalyLMM-style verifier, and an
SSDC-inspired structured verifier. Their calibrated scores are then combined by
static late fusion and an uncertainty-gated consensus module. Our experts are
non-generative rerankers already scored on the shared pool, so debate costs no
additional forward pass; we gate instead for reliability, since debate can
overturn an initially correct prediction in $3.4$--$5.7\%$ of visual-question-answering
queries~\cite{fan2026imad}. Our consensus is a credibility-weighted logarithmic
opinion pool~\cite{degroot1974consensus,genest1986combining} over expert beliefs
rather than an exchange of natural-language arguments.

\section{Proposed Method}
\label{sec:method}

\begin{figure*}[!t]
	\centering
	\includegraphics[width=\textwidth]{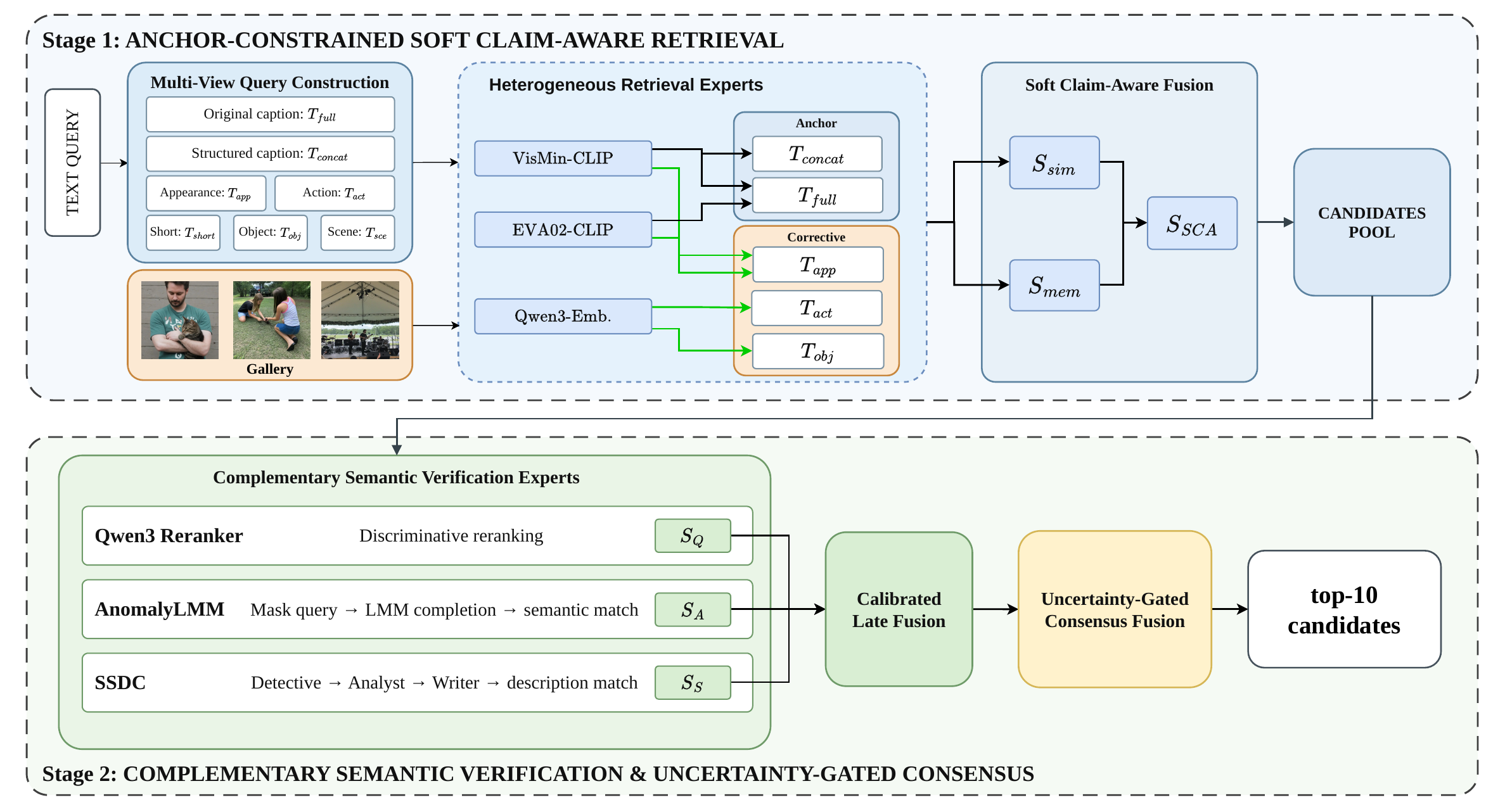}
	\vspace{-0.5cm}
	\caption{\textbf{Overview of the our framework.}
Anchor-constrained multi-view retrieval constructs a shared top-$K$ candidate pool. Three rerankers refine the candidates, and uncertainty-gated consensus combines their scores to produce the final ranking.}
	\label{fig:pipeline}
	\vspace{-0.5cm}
\end{figure*}

Fig.~\ref{fig:pipeline} summarizes the proposed two-stage framework: gallery-wide
multi-view retrieval produces a shared top-$10$ pool, which three parallel
rerankers score before an uncertainty-gated consensus produces the final ranking.

\subsection{Multi-View Query Construction}
\label{sec:query_views}

\begin{figure*}[!t]
	\centering
	\includegraphics[width=\textwidth]{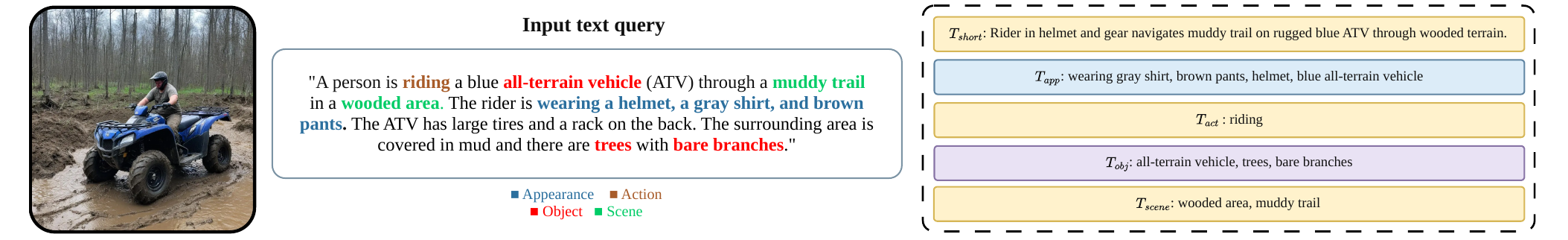}
	\vspace{-0.5cm}
	\caption{The input description is decomposed into retrieval-oriented appearance, action, object, and scene facets.}
	\label{fig:decompose}
	\vspace{-0.75cm}
\end{figure*}

Given a query $q$, we construct a full description
$T_{\mathrm{full}}$ and several semantic views describing appearance,
action, object interaction, and scene. We also form a structured caption
\begin{equation}
T_{\mathrm{concat}}
=
T_{\mathrm{short}}
\oplus T_{\mathrm{app}}
\oplus T_{\mathrm{act}}
\oplus T_{\mathrm{obj}}
\oplus T_{\mathrm{scene}},
\label{eq:tconcat}
\end{equation}
where $\oplus$ denotes string concatenation. The full and concatenated
captions preserve global semantics, while individual facets provide
fine-grained evidence (Fig.~\ref{fig:decompose}). Since decomposed facets may be incomplete or ambiguous, they are used only as soft corrective signals rather than independent retrieval constraints.

\subsection{Anchor-Constrained Soft Claim-Aware Retrieval}
\label{sec:sca}

We combine heterogeneous vision-language retrievers operating on
different textual views. Each branch $b$ consists of a retrieval model,
a textual view $T_b$, and a weight $w_b$. The image-text similarity is
\begin{equation}
s_{b,i}
=
\operatorname{cos}
\left(
f_b^{T}(T_b),
f_b^{I}(I_i)
\right).
\label{eq:branch_score}
\end{equation}

Branches using $T_{\mathrm{full}}$ or $T_{\mathrm{concat}}$ are treated
as \emph{anchors}, whereas appearance, action, and object branches act
as low-weight corrective experts.

\subsubsection{Anchor-Constrained Candidate Generation}

Let $\mathcal{R}_b(q)$ denote the candidate list retrieved by branch
$b$. We first take the union of all branch outputs and retain only
candidates supported by at least one anchor branch:
\begin{equation}
\mathcal{C}_{\mathrm{anc}}(q)
= \left(\bigcup_{b\in\mathcal{B}}\mathcal{R}_b(q)\right)
\cap
\left(\bigcup_{b\in\mathcal{B}_{\mathrm{anc}}}\mathcal{R}_b(q)\right).
\label{eq:anchor_constraint}
\end{equation}
This constraint prevents an isolated and potentially noisy facet from
introducing candidates that are inconsistent with the complete query.

\subsubsection{Robust Score Calibration}

Because different retrievers produce scores on different scales, each
branch is normalized independently for every query:
\begin{equation}
\widehat{s}_{b,i}
=
\operatorname{clip}_{[-3,3]}
\left(
\frac{
s_{b,i}-\operatorname{median}(\mathbf{s}_b)
}{
\max\left(
Q_{0.95}(\mathbf{s}_b)-Q_{0.05}(\mathbf{s}_b),
\epsilon
\right)
}
\right),
\label{eq:robust_norm}
\end{equation}
where $\mathbf{s}_b$ denotes the retrieved scores of branch $b$.
Percentile-based scaling reduces sensitivity to outliers, while clipping
limits the influence of extreme values.

\subsubsection{Soft Claim-Aware Fusion}

The calibrated scores are combined using a fixed denominator:
\begin{equation}
S_{\mathrm{sim}}(q,I_i)
=
\frac{
\sum_{b\in\mathcal{B}}w_b\widehat{s}_{b,i}
}{
\sum_{b\in\mathcal{B}}w_b
},
\label{eq:fixed_denom_fusion}
\end{equation}
where $\widehat{s}_{b,i}=0$ when candidate $I_i$ is absent from branch
$b$. This formulation avoids overestimating candidates supported by
only a small number of branches.

We further introduce a soft agreement term based on reciprocal ranks:
\begin{equation}
S_{\mathrm{mem}}(q,I_i)
=
\frac{
\sum_{b\in\mathcal{B}}
w_b\,
\mathbb{I}[I_i\in\mathcal{R}_b(q)]
/r_{b,i}
}{
\sum_{b\in\mathcal{B}}w_b
},
\label{eq:membership_score}
\end{equation}
where $r_{b,i}$ is the rank of candidate $I_i$ in branch $b$. The final
retrieval score is
\begin{equation}
S_{\mathrm{SCA}}(q,I_i)
=
S_{\mathrm{sim}}(q,I_i)
+
\lambda_{\mathrm{mem}}S_{\mathrm{mem}}(q,I_i).
\label{eq:sca_score}
\end{equation}
The highest-ranked candidates are passed to the subsequent semantic
reranking modules.

\subsection{PAB-Adapted Qwen3 Reranking}
\label{sec:qwen_reranker}

Embedding retrieval efficiently narrows the search space, but its independent
image and text representations provide limited cross-modal interaction.

We therefore employ a Qwen3-based multimodal reranker \cite{qwen3vltechnical} to independently
evaluate every query-candidate pair in the shared pool
$\mathcal{C}_{10}(q)$:
\begin{equation}
S_{Q}(q,I_i)
=
R_{\phi}(q,I_i),
\qquad
I_i\in\mathcal{C}_{10}(q),
\label{eq:qwen_score}
\end{equation}
where $R_{\phi}$ outputs a fine-grained relevance score.

To adapt the reranker to both natural test queries and structured semantic
descriptions, we fine-tune it using a mixed-caption strategy. For every
training image, its textual input is sampled as
\begin{equation}
T_{\mathrm{train}}
=
\begin{cases}
T_{\mathrm{full}}, & \text{with probability }0.5,\\
T_{\mathrm{concat}}, & \text{with probability }0.5.
\end{cases}
\label{eq:mixed_caption}
\end{equation}
Training with $T_{\mathrm{full}}$ preserves natural language composition and
relations among concepts, whereas $T_{\mathrm{concat}}$ encourages the model
to explicitly attend to appearance, action, object, and scene evidence. The
equal mixture reduces overfitting to either a single caption template or a
specific decomposition format.

At inference, the reranker scores all $10$ candidates selected by
Eq.~\eqref{eq:sca_score}, making deeper image-text interaction
computationally feasible. Its independent ranking is
\begin{equation}
\mathcal{R}_{Q}(q)
=
\operatorname{Sort}_{I_i\in\mathcal{C}_{10}(q)}
S_Q(q,I_i).
\label{eq:qwen_ranking}
\end{equation}

\subsection{Complementary Semantic Verification}
\label{sec:semantic_verification}

Qwen3 \cite{qwen3technical}, AnomalyLMM, and SSDC operate as three parallel reranking experts on the same candidate support. In particular, the two semantic verification branches do not consume a candidate list filtered by Qwen3; each branch independently assigns a score to every
$I_i\in\mathcal{C}_{10}(q)$. This common support makes their calibrated scores directly comparable and allows inter-expert disagreement to be measured in the subsequent consensus module.

\subsubsection{AnomalyLMM-Style Cloze Verification}

Following the masked cross-modal reasoning strategy of
AnomalyLMM~\cite{ju2025anomalylmm}, we replace visually discriminative words,
such as action verbs and color attributes, with structured placeholders:
\begin{equation}
q_{\mathrm{mask}}
=
\operatorname{Mask}(q).
\end{equation}
For each candidate $I_i\in\mathcal{C}_{10}(q)$, an LMM reconstructs the
masked information from visual evidence:
\begin{equation}
c_i^{A}
=
\operatorname{LMM}
\left(
I_i,q_{\mathrm{mask}}
\right),
\qquad
I_i\in\mathcal{C}_{10}(q).
\end{equation}
\begin{equation}
S_A(q,I_i)
=
\operatorname{SemMatch}
\left(
q,c_i^{A}
\right).
\end{equation}

The resulting scores induce the independent ranking
\begin{equation}
\mathcal{R}_{A}(q)
=
\operatorname{Sort}_{I_i\in\mathcal{C}_{10}(q)}
S_A(q,I_i).
\end{equation}

By forcing the model to infer missing visual attributes from the candidate image, this branch provides explicit verification of action and attribute consistency beyond global image-text similarity.

\subsubsection{SSDC-Inspired Structured Verification}

We additionally adapt the semantic verification stage of
SSDC~\cite{xie2026ssdc}, which decomposes multimodal reasoning into three
specialized roles. A Detective first filters obvious mismatches, an Analyst
extracts structured visual evidence, and a Writer synthesizes the evidence into
a candidate-specific description:
\begin{align}
d_i &=
\operatorname{Detective}(q,I_i),\\
e_i &=
\operatorname{Analyst}(q,I_i,d_i),\\
c_i^{S} &=
\operatorname{Writer}(e_i).
\end{align}
The SSDC semantic score is then obtained by measuring the correspondence
between the original query and the synthesized description:
\begin{equation}
S_S(q,I_i)
=
\operatorname{sim}
\left(
E_T(q),
E_T(c_i^{S})
\right),
\end{equation}
where $E_T$ denotes the text embedding model. Here $E_T$ is the Qwen3-Embedding-8B encoder of Table~\ref{tab:implementation_config}, reused to score text-text similarity, and $\operatorname{SemMatch}(a,b)=\cos(E_T(a),E_T(b))$. Compared with the AnomalyLMM-style branch, which focuses on explicit completion of critical attributes and actions, the SSDC-inspired branch captures broader semantic evidence including pose, appearance, object interaction, and scene context. The two branches therefore provide complementary verification signals for resolving difficult candidates.

\subsection{Calibrated Late Fusion}
\label{sec:late_fusion}

The three rerankers score the same $K=10$ candidates but produce values on different scales. We therefore normalize each expert $m\in\mathcal{M}:=\{Q,A,S\}$ independently for every query:
\begin{equation}
\overline{S}_{m,i}
=
\operatorname{clip}_{[-3,3]}
\left(
\frac{
S_m(q,I_i)-\operatorname{median}_{j} S_m(q,I_j)
}{
\operatorname{IQR}_{j} S_m(q,I_j)+\epsilon
}
\right),
\label{eq:late_score_calibration}
\end{equation}
where $I_j\in\mathcal{C}_{10}(q)$ and $\epsilon>0$. Using $\boldsymbol{\omega}=(\omega_Q,\omega_A,\omega_S)=(0.3,0.3,0.4)$, the fused score is
\begin{equation}
S_{\mathrm{LF},i}
=
\sum_{m\in\mathcal{M}}
\omega_m\overline{S}_{m,i}.
\label{eq:late_fusion_score}
\end{equation}
We convert it into a belief over the shared candidate pool:
\begin{equation}
\pi^{\mathrm{LF}}_i
=
\frac{
\exp(S_{\mathrm{LF},i}/\tau_{\mathrm{LF}})
}{
\sum_{j=1}^{K}
\exp(S_{\mathrm{LF},j}/\tau_{\mathrm{LF}})
}.
\label{eq:late_fusion_belief}
\end{equation}
This distribution serves as both the static-fusion baseline and the reference for estimating query uncertainty. It is not used as the consensus prior because it already includes all three expert outputs.

\vspace{-0.5cm}\subsection{Uncertainty-Gated Consensus Fusion}
\label{sec:debate_fusion}
\vspace{-0.25cm}

\begin{figure*}[!t]
	\centering
	\includegraphics[width=\textwidth]{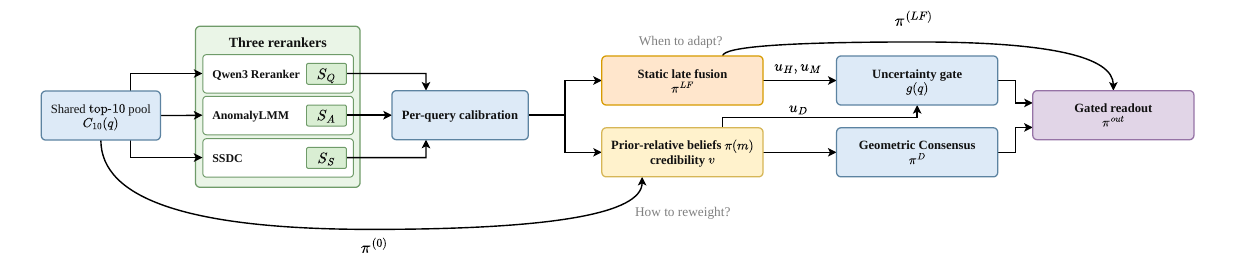}
	\vspace{-0.5cm}
	\caption{\textbf{Uncertainty-gated consensus.} Calibrated reranker scores are adaptively combined using uncertainty, disagreement, and expert credibility, reverting to standard late fusion when $g(q)=0$.}
    \label{fig:consensus_module}
	\vspace{-0.75cm}
\end{figure*}

Static fusion applies fixed expert weights to every query. We instead activate an adaptive consensus when the fused prediction is uncertain. The module uses only the scores already computed on $\mathcal{C}_{10}(q)$ and requires no additional multimodal forward pass (Fig.~\ref{fig:consensus_module}).

\paragraph{Shared prior and expert beliefs.}
The upstream SCA scores provide a common prior independent of the rerankers. Let $S_{0,i}=S_{\mathrm{SCA}}(q,I_i)$. We normalize them as
\begin{equation}
\overline{S}_{0,i}
=
\operatorname{clip}_{[-3,3]}
\left(
\frac{
S_{0,i}
-
\operatorname{median}_{j} S_{0,j}
}{
\operatorname{IQR}_{j} S_{0,j}
+
\epsilon
}
\right),
\qquad
I_j\in\mathcal{C}_{10}(q),
\label{eq:sca_prior_calibration}
\end{equation}
and define
\begin{equation}
\pi_i^{(0)}
=
\frac{
\exp(\overline{S}_{0,i}/\tau_0)
}{
\sum_{j=1}^{K}
\exp(\overline{S}_{0,j}/\tau_0)
},
\qquad i=1,\ldots,K.
\label{eq:debate_prior}
\end{equation}
Each expert updates this prior using its calibrated evidence:
\begin{equation}
\pi_i^{(m)}
=
\frac{
\pi_i^{(0)}\exp(\overline{S}_{m,i}/\tau_m)
}{
\sum_{j=1}^{K}\pi_j^{(0)}\exp(\overline{S}_{m,j}/\tau_m)
},
\qquad m\in\mathcal{M},
\label{eq:expert_belief}
\end{equation}
where $\tau_m>0$. The prior-relative evidence is $t_i^{(m)}=\log(\pi_i^{(m)}/\pi_i^{(0)})$. We use Shannon entropy $H(\cdot)$ and Jensen-Shannon divergence $\operatorname{JS}(\cdot,\cdot)$ \cite{lin1991divergence}.

\paragraph{Uncertainty gate.}
Let $\pi_{(1)}^{\mathrm{LF}}\geq\pi_{(2)}^{\mathrm{LF}}$ be the two largest late-fusion probabilities. We measure uncertainty using entropy, top-two margin, and pairwise expert disagreement:
\begin{align}
u_H(q)
&=
\frac{H(\pi^{\mathrm{LF}})}{\log K},
\label{eq:entropy_uncertainty}\\
u_M(q)
&=
1-
\left(
\pi_{(1)}^{\mathrm{LF}}
-
\pi_{(2)}^{\mathrm{LF}}
\right),
\label{eq:margin_uncertainty}\\
u_D(q)
&=
\frac{1}{3\log 2}
\sum_{\substack{m<n\\m,n\in\mathcal{M}}}
\operatorname{JS}
\left(
\pi^{(m)},\pi^{(n)}
\right).
\label{eq:disagreement_uncertainty}
\end{align}
Their average $u(q)=(u_H(q)+u_M(q)+u_D(q))/3$ determines the continuous gate
\begin{equation}
g(q)
=
\operatorname{clip}_{[0,1]}
\left(
\frac{
u(q)-\delta_{\mathrm{low}}
}{
\delta_{\mathrm{high}}-\delta_{\mathrm{low}}
}
\right).
\label{eq:debate_gate}
\end{equation}
Thus, $g(q)=0$ below $\delta_{\mathrm{low}}$, $g(q)=1$ above $\delta_{\mathrm{high}}$, and intermediate values yield proportional interpolation.

\paragraph{Adaptive credibility and consensus.}
Let $D_{mn}=\operatorname{JS}(\pi^{(m)},\pi^{(n)})$, $\bar{d}_m=\frac{\sum_{n\neq m}D_{mn}}{2}$. We reduce the credibility of an expert that is isolated from the others:
\begin{equation}
v_m
=
\frac{
\omega_m\exp\left(-\bar{d}_m/\theta_{\mathrm{JS}}\right)
}{
\sum_{\ell\in\mathcal{M}}
\omega_\ell\exp\left(-\bar{d}_\ell/\theta_{\mathrm{JS}}\right)
},
\qquad
\sum_{m\in\mathcal{M}}v_m=1.
\label{eq:debate_trust}
\end{equation}
When the experts agree, $\boldsymbol{v}=\boldsymbol{\omega}$; when one expert is isolated, its credibility is reduced. The resulting logarithmic opinion pool~\cite{degroot1974consensus,genest1986combining,cooke1991experts} is
\begin{equation}
\pi_i^{D}
=
\frac{
\prod_{m\in\mathcal{M}}\left(\pi_i^{(m)}\right)^{v_m}
}{
\sum_{j=1}^{K}\prod_{m\in\mathcal{M}}\left(\pi_j^{(m)}\right)^{v_m}
}
\;\propto\;
\pi_i^{(0)}
\exp\left(
\sum_{m\in\mathcal{M}}v_m t_i^{(m)}
\right).
\label{eq:debate_consensus}
\end{equation}
Because $\sum_m v_m=1$, the shared prior is counted once, while candidates supported mainly by an isolated expert are suppressed.

\paragraph{Gated output.}
The final belief interpolates between static fusion and adaptive consensus:
\begin{equation}
\pi_i^{\mathrm{out}}
=
\frac{
\left(\pi_i^{\mathrm{LF}}\right)^{1-g(q)}
\left(\pi_i^{D}\right)^{g(q)}
}{
\sum_{j=1}^{K}
\left(\pi_j^{\mathrm{LF}}\right)^{1-g(q)}
\left(\pi_j^{D}\right)^{g(q)}
},
\qquad
\mathcal{R}_{\mathrm{final}}(q)
=
\operatorname{Sort}_{I_i\in\mathcal{C}_{10}(q)}
\pi_i^{\mathrm{out}}.
\label{eq:gated_output}
\end{equation}
Hence, $g(q)=0$ exactly recovers calibrated late fusion, while $g(q)=1$ uses the full consensus. Equivalently,
\begin{equation}
\pi_i^{\mathrm{out}}
\;\propto\;
\exp\left[
\frac{g(q)}{\tau_0}\,\overline{S}_{0,i}
+
\sum_{m\in\mathcal{M}}
\underbrace{\left(
\frac{(1-g(q))\,\omega_m}{\tau_{\mathrm{LF}}}
+
\frac{g(q)\,v_m(q)}{\tau_m}
\right)}_{\alpha_m(q)}
\,\overline{S}_{m,i}
\right].
\label{eq:debate_closed_form}
\end{equation}
The gate therefore adjusts the expert weights and reintroduces the SCA prior only for uncertain queries. Its computational cost is $\mathcal{O}(|\mathcal{M}|^2K)$, which is negligible for $|\mathcal{M}|=3$ and $K=10$.

All temperatures, $\theta_{\mathrm{JS}}$, $\delta_{\mathrm{low}}$, and $\delta_{\mathrm{high}}$ are selected using only the held-out synthetic validation set. No test-side calibration is performed, although applying these thresholds to real queries may remain sensitive to Sim2Real calibration drift~\cite{fan2026imad,verma2026selene}.

\begin{table}
\centering
\caption{Implementation configs of the retrieval and semantic reranking modules.}
\vspace{-0.25cm}
\label{tab:implementation_config}
\small
\setlength{\tabcolsep}{3.5pt}
\renewcommand{\arraystretch}{1.08}

\resizebox{\linewidth}{!}{%
\begin{tabular}{@{}lccc@{\hspace{1.3em}}ll@{}}
\toprule
\multicolumn{4}{c}{\textbf{Retrieval branches}}
&
\multicolumn{2}{c}{\textbf{Semantic reranking models}}
\\
\cmidrule(lr){1-4}
\cmidrule(lr){5-6}

Backbone and view
& Top-$K$
& Weight
& Role
& Stage or role
& Model
\\
\midrule

VisMin-CLIP, $T_{\mathrm{concat}}$
& 200 & 0.350 & Anchor
& AnomalyLMM masking
& Qwen3-14B
\\

VisMin-CLIP, $T_{\mathrm{full}}$
& 200 & 0.250 & Anchor
& AnomalyLMM completion
& Qwen3-VL-8B
\\

Qwen3-VL-Emb., $T_{\mathrm{full}}$
& 200 & 0.300 & Anchor
& SSDC Detective
& Qwen3-VL-8B-Inst.
\\

\cmidrule(lr){1-4}
VisMin-CLIP, $T_{\mathrm{app}}$
& 200 & 0.030 & Corrective
& SSDC Analyst
& Qwen3-VL-8B-Inst.
\\

EVA02-CLIP, $T_{\mathrm{app}}$
& 200 & 0.020 & Corrective
& SSDC Writer
& Qwen3-VL-8B-Inst.
\\

Qwen3-VL-Emb., $T_{\mathrm{act}}$
& 100 & 0.006 & Corrective
& Qwen3 reranker
& Qwen3-VL-Reranker-8B
\\

Qwen3-VL-Emb., $T_{\mathrm{obj}}$
& 100 & 0.006 & Corrective
& &
\\

\bottomrule
\end{tabular}%
}

\vspace{2pt}
\begin{minipage}{\linewidth}
\footnotesize
\textit{Note:} Qwen3-VL-Emb. denotes Qwen3-VL-Embedding-8B;
Qwen3-VL-8B-Inst. denotes Qwen3-VL-8B-Instruct; and
Qwen3-VL-Reranker-8B denotes Qwen3-VL-Reranker-8B.
\end{minipage}

\vspace{-0.5cm}
\end{table}

\vspace{-0.5cm}
\section{Experiments}
\vspace{-0.25cm}

\label{sec:exp}
\subsection{Dataset and Evaluation}




We evaluate our method on the PAB benchmark for Sim2Real text-based
person anomaly search~\cite{yang2024beyond}. The original training set
contains $1{,}013{,}605$ synthetic image--text pairs. We randomly hold
out $2{,}000$ samples as a validation set and use the remaining samples
for model training. The official test set contains $1{,}978$ textual
queries and a gallery of $1{,}978$ ground-truth images together with
$34{,}795$ distractors.

The validation split is constructed exclusively from the synthetic
training data. All architectural choices, branch and fusion weights,
temperatures, gate thresholds, and other hyperparameters are selected
exclusively on this validation split. No official test query, gallery
image, evaluation score, or other test-side output is used for model
selection, calibration, or post-processing adjustment.

Tables~\ref{tab:component_ablation}
and~\ref{tab:retrieval_ablation} report official scores returned by the
AI City~2026 Track~4 evaluation server for a predefined set of frozen
ablation configurations. We used 20 of the permitted submission
attempts under the official evaluation protocol. Each configuration was fully specified from the training and validation data before its server evaluation, and the returned test scores were used only for reporting the corresponding ablation result. We report mAP@10, Recall@1 (R@1), and Recall@5 (R@5), with all results expressed as percentages. 

\subsection{Implementation Details}
\label{sec:implementation}

\paragraph{Facet decomposition.}
The facet views of Sec.~\ref{sec:query_views} are produced by an LLM
decomposition procedure that instantiates CREDENCE~\cite{tran2026credence}. Each
caption is first split into five typed, independently retrievable sub-queries of
at most twenty words: \emph{appearance} (garment, color, and body attributes,
keeping bindings such as ``black jacket'' intact), \emph{action}, \emph{objects}
(non-person items), \emph{scene} (location and environment only), and a
pronoun-free short summary $T_{\mathrm{short}}$. A reflection pass then self-corrects the decomposition
against the four CREDENCE criteria, atomicity (one assertion per sub-query),
fidelity (the sub-queries jointly cover the caption), entity preservation (every
color, object, count, and location survives in some sub-query), and
non-redundancy, while enforcing the scene/object separation. The same
procedure is applied offline to all $1{,}013{,}605$ training captions and online
to each test query; the resulting facets populate the retrieval branches of
Table~\ref{tab:implementation_config}.

The decomposition and reflection passes are performed by
Llama-3.3-70B-Instruct (4-bit) for test queries and Llama-3-8B-Instruct for the
training captions, with near-greedy decoding (temperature~$0.05$) \cite{llama3herd}.

The retrieval ensemble contains a VisMin-initialized CLIP, a PAB-adapted EVA02-CLIP, and the zero-shot Qwen/Qwen3-VL-Embedding-8B multimodal embedding model \cite{qwen3vlembedding}. Following the model's standard inference procedure, the gallery images and textual query views are represented in a shared embedding space and compared using cosine similarity.
For the CLIP and EVA02 branches, only the final three Transformer
blocks of the image and text encoders are fine-tuned.

The VisMin-initialized CLIP is trained using $T_{\mathrm{full}}$, $T_{\mathrm{concat}}$, and $T_{\mathrm{short}}$ with sampling probabilities $0.5$, $0.4$, and $0.1$, respectively. The retrieval and semantic verification configurations are summarized in Table~\ref{tab:implementation_config}.

For AnomalyLMM, Qwen3-14B constructs the masked query, while Qwen3-VL-8B completes the masked attributes using candidate-image evidence. For SSDC, the Detective, Analyst, and Writer agents all use Qwen3-VL-8B-Instruct. Qwen3-Embedding-8B is used in the retrieval stage and, in Stage~2, only as the text-text encoder $E_T$ of the semantic-verification branches.

The Qwen3 reranker is a Qwen3-VL-8B backbone fine-tuned as a pointwise
(yes/no) relevance scorer with LoRA (rank~$16$, $\alpha{=}32$; the vision
encoder frozen) \cite{hu2022lora} for one epoch at learning rate $1\times10^{-4}$ on approximately $30$k synthetic PAB queries, using the mixed-caption sampling of Eq.~\eqref{eq:mixed_caption} ($T_{\mathrm{full}}/T_{\mathrm{concat}}$ at
$0.5/0.5$) with seven hard negatives per query mined by embedding
nearest-neighbor search.

For the consensus module we use $\tau_0=\tau_{\mathrm{LF}}=\tau_Q=\tau_A=\tau_S=1.0$
and $\theta_{\mathrm{JS}}=0.5$; the gate thresholds $\delta_{\mathrm{low}}$ and
$\delta_{\mathrm{high}}$ are set to the $30$th and $90$th percentiles of the
validation $u(q)$ distribution.

We set $\lambda_{\mathrm{mem}}=0.05$. The top-10 candidates ranked by $S_{\mathrm{SCA}}$ form the shared input pool for the Qwen3 reranker, AnomalyLMM, and SSDC.

\vspace{-0.5cm}
\section{Results and Discussion}
\vspace{-0.25cm}

\subsection{Component Ablation}



Table~\ref{tab:component_ablation} reports the official-test results
of the predefined reranking ablations, all evaluated using the same
soft claim-aware retrieval baseline. Each individual reranker improves
mAP@10 and R@1. Qwen3 and AnomalyLMM each add approximately $4.1$
mAP points, while SSDC provides the largest individual improvements in
mAP@10 and R@1, with a small R@5 trade-off.

Combining the three verification procedures provides a substantially
stronger result than using any individual reranker. The
validation-selected weighted fusion reaches $95.18\%$ mAP@10,
$92.47\%$ R@1, and $98.48\%$ R@5 on the official test set.
The full uncertainty-gated consensus further reaches $95.41\%$
mAP@10, $94.44\%$ R@1, and $99.09\%$ R@5. These official-test
ablations indicate that adaptive score-level consensus is beneficial
over the corresponding fixed fusion configuration.

\vspace{-0.5cm}
\begin{figure*}
	\centering
	\includegraphics[width=\textwidth]{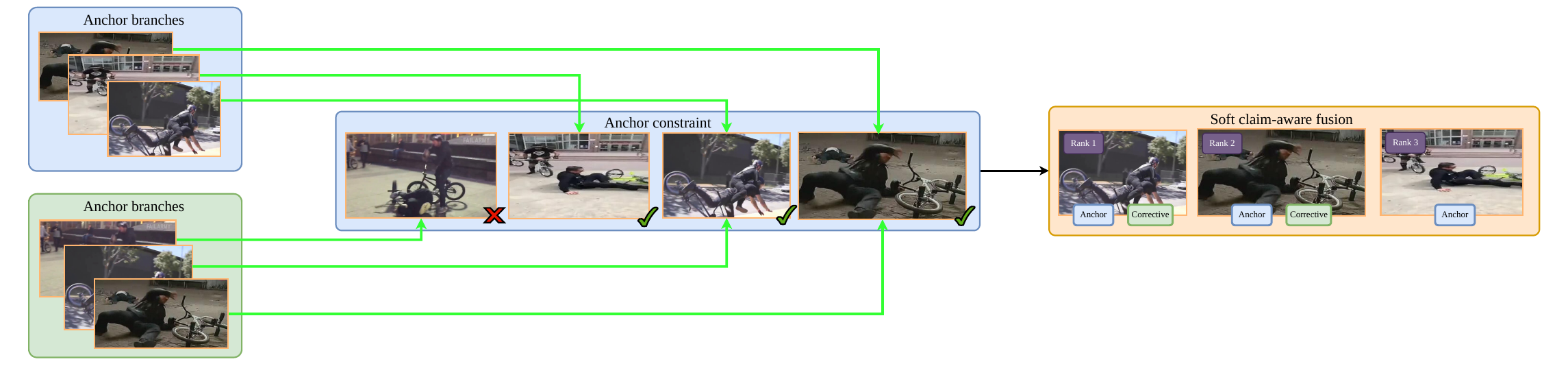}
	\vspace{-0.5cm}
	\caption{Anchor branches define the valid candidate pool, while corrective branches only refine the scores of anchor-supported candidates through soft claim-aware fusion.}
    \label{fig:sca}
	\vspace{-0.75cm}
\end{figure*}
\subsection{Retriever and Anchor Analysis}

Table~\ref{tab:retrieval_ablation} evaluates the retrieval branches and anchor design. OpenCLIP-VisMin is the strongest standalone retriever in mAP@10 and R@1, while EVA02 attains the highest R@5, confirming their complementary behavior.

Anchoring improves performance, with the VM-concat anchor adding about $1.5$ mAP points and multiple anchors yielding further gains. The full soft claim-aware configuration reaches $86.4407$ mAP@10, outperforming the best standalone retriever by $8.92$ points. These results support using global captions as anchors and fine-grained facets as bounded corrective cues.

\vspace{-0.5cm}
\begin{table*}
\centering
\begin{minipage}[t]{0.49\textwidth}
\centering
\scriptsize
\captionof{table}{Official-test ablation of the reranking components.
All configurations and fusion weights were selected using only the
synthetic validation split. $(0.3, 0.3, 0.4)$ are weights for Qwen3, AnomalyLMM, and SSDC.}
\vspace{-0.25cm}
\label{tab:component_ablation}
\setlength{\tabcolsep}{3pt}
\renewcommand{\arraystretch}{1.05}
\resizebox{\linewidth}{!}{
\begin{tabular}{lccc}
\toprule
Configuration & mAP@10 & R@1 & R@5 \\
\midrule
Soft claim-aware retrieval
    & 86.4407 & 78.2103 & 96.2588 \\
\quad + Qwen3 reranker
    & 90.5258 & 84.4287 & 97.9272 \\
\quad + AnomalyLMM
    & 90.5629 & 84.4287 & 97.9272 \\
\quad + SSDC
    & 91.3145 & 85.8443 & 97.8261 \\
\quad + Qwen3 + SSDC
    & 91.7792 & 86.6532 & 98.0283 \\
\quad + all, fusion $(0.3,0.3,0.4)$
    & 95.1782 & 92.4671 & 98.4833 \\
\quad + all, consensus
    & \textbf{95.4078}
    & \textbf{94.4388}
    & \textbf{99.0900} \\
\bottomrule
\end{tabular}
}
\end{minipage}
\hfill
\begin{minipage}[t]{0.49\textwidth}
\centering
\scriptsize
\captionof{table}{Official-test ablation of the retrieval branches and
anchor configurations. All branch choices and weights were fixed using
only the synthetic validation split. VM denotes the VisMin-pretrained OpenCLIP branch.}
\vspace{-0.25cm}
\label{tab:retrieval_ablation}
\setlength{\tabcolsep}{3pt}
\renewcommand{\arraystretch}{1.05}
\resizebox{\linewidth}{!}{
\begin{tabular}{lccc}
\toprule
Retrieval configuration & mAP@10 & R@1 & R@5 \\
\midrule
EVA02 only
    & 73.1893 & 58.8473 & 92.3660 \\
Qwen3-VL-Embedding only
    & 76.8495 & 66.3802 & 89.2821 \\
OpenCLIP-VisMin only
    & 77.5226 & 66.8857 & 91.0010 \\
No anchor
    & 82.8796 & 73.3064 & 94.3377 \\
VM-concat anchor, no membership
    & 84.3439 & 75.2275 & 95.2983 \\
VM-concat anchor
    & 84.3932 & 75.1769 & 95.4499 \\
Multiple anchors, no membership
    & 84.6248 & 75.0253 & 96.0061 \\
Main soft claim-aware configuration
    & \textbf{86.4407}
    & \textbf{78.2103}
    & \textbf{96.2588} \\
\bottomrule
\end{tabular}
}
\end{minipage}

\end{table*}

\vspace{-1cm}
\subsection{Fusion Strategy}
\label{sec:fusion_strategy}


The fixed fusion weights $(0.3,0.3,0.4)$ for Qwen3, AnomalyLMM,
and SSDC were selected exclusively on the synthetic validation split and frozen before official evaluation. This configuration achieves $95.18\%$ mAP@10 on the official test set. As a predefined diagnostic ablation, removing Qwen3 and using weights $(0,0.5,0.5)$ yields $93.4783\%$ R@1 but reduces R@5 to $93.6299\%$. This result suggests that the semantic verifiers can strongly promote a top candidate while producing a less stable top-$k$ ranking, whereas the discriminative Qwen3 reranker provides a stabilizing signal.

Per-query median--IQR calibration makes the expert scores comparable before fusion. In particular, it is approximately invariant to expert-specific positive shifts and rescaling, up to clipping and the stabilizing term $\epsilon$. Moreover, because the calibrated scores are clipped to $[-3,3]$ and the fusion weights are non-negative with $\sum_m\omega_m=1$, the fused score is also bounded within this interval. These properties reduce the risk that an expert dominates only because of its numerical score range. Nevertheless, calibrated late fusion remains static: the same weights are applied whether the experts agree or strongly disagree.

Selective collaboration has previously been explored in SELENE, DOWN, and iMAD, which activate multi-agent debate according to disagreement, confidence, or learned hesitation signals~\cite{verma2026selene,eo2025down,fan2026imad}. Our setting follows the same general principle of adapting only when needed, but differs in that all three rerankers have already scored the same candidate pool. The gate therefore controls score-level consensus rather than triggering additional generative interaction, introducing no extra multimodal forward pass.

\begin{figure}
\centering

\begin{minipage}[t]{0.40\linewidth}
    \vspace{0pt}
    \centering

    \includegraphics[width=\linewidth]
    {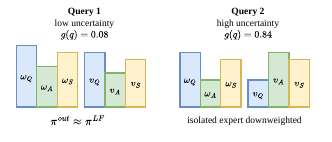}

    \vspace{-0.5cm}

    \caption{\textbf{Query-adaptive consensus weighting.} Low uncertainty queries keep $g(q)$ small and $\bm{v}\!\approx\!\bm{\omega}$, high uncertainty queries increase $g(q)$ and downweight isolated experts.}
    \label{fig:consensus_weight_shift}
\end{minipage}
\hfill
\begin{minipage}[t]{0.59\linewidth}
    \vspace{0pt}
    \centering

    \captionof{table}{Main design choices selected on the held-out synthetic validation set. Only the selected configuration was evaluated on the official test server.}
    \label{tab:design_choices}

    \vspace{1mm}
    \scriptsize
    \setlength{\tabcolsep}{3pt}
    \renewcommand{\arraystretch}{1.05}

    \begin{tabular}{@{}lcc@{}}
        \toprule
        \textbf{Component}
        & \textbf{Alternatives}
        & \textbf{Selected} \\
        \midrule
        Anchor strategy
        & None / Single / Multiple
        & \textbf{Multiple} \\

        Candidate pool
        & Top-5 / Top-10 / Top-20
        & \textbf{Top-10} \\

        Membership term
        & Off / On
        & \textbf{On} \\

        Fusion weights
        & Grid search
        & \textbf{(0.3,0.3,0.4)} \\

        Consensus fusion
        & Static / Adaptive
        & \textbf{Adaptive} \\
        \bottomrule
    \end{tabular}
\end{minipage}

\end{figure}

The module separates \emph{when to adapt} from \emph{how to reweight}. Entropy, top-two margin, and pairwise Jensen-Shannon disagreement determine the gate $g(q)$, while query-specific credibilities reduce the influence of isolated experts. As illustrated in Fig.~\ref{fig:consensus_weight_shift}, low-uncertainty queries remain close to calibrated late fusion, whereas ambiguous queries shift progressively toward the adaptive consensus.

The geometric consensus also has a variational interpretation:
\begin{equation}
\pi^{D}
=
\arg\min_{\pi}
\sum_{m\in\mathcal{M}}
v_m\operatorname{KL}
\left(
\pi\,\|\,\pi^{(m)}
\right).
\end{equation}
Hence, $\pi^{D}$ is the distribution closest to the expert beliefs under their query-dependent credibilities. Because $\sum_m v_m=1$, the shared SCA prior is counted exactly once, while the prior-relative evidence of the rerankers is combined. Compared with arithmetic averaging, the geometric pool also suppresses candidates that receive support mainly from an isolated expert.

On the official test set, adaptive consensus improves the validation-selected weighted fusion from $95.18\%$ to $95.41\%$ mAP@10, while increasing R@1 and R@5 by $1.97$ and $0.61$ percentage points, respectively. Since the evaluation server returns only aggregate metrics, this difference should be interpreted as a net improvement rather than as an identification of the specific queries corrected by consensus. The results are consistent with selective, query-dependent conflict resolution being preferable to applying adaptive weights to every query. However, agreement does not necessarily imply correctness: correlated experts may agree on an incorrect candidate. The exact late-fusion fallback limits this risk by applying adaptive consensus only when the uncertainty signals exceed the validation-calibrated thresholds.


\vspace{-0.5cm}

\vspace{-0.5cm}

\vspace{-0.75cm}
\section{Conclusion}
\vspace{-0.25cm}
We presented a coarse-to-fine framework for Sim2Real text-based person anomaly search, combining anchor-constrained multi-view retrieval with three complementary rerankers and uncertainty-gated consensus. Global captions preserve candidate recall, while decomposed facets provide bounded corrective evidence.

Experiments on PAB show that retrieval fusion and semantic reranking could improve performance, with the consensus-based configuration achieving the best results. These findings support restricting expensive multimodal reasoning to a small candidate pool and adapting expert fusion only for uncertain queries. Future work will explore more efficient verifier selection and adaptive routing.



%
%
\bibliographystyle{splncs04}
\bibliography{main}

@String(CVPR  = {IEEE Conf. Comput. Vis. Pattern Recog.})

@String(ICCV  = {Int. Conf. Comput. Vis.})

@String(ECCV  = {Eur. Conf. Comput. Vis.})

@String(NeurIPS = {Adv. Neural Inform. Process. Syst.})

@String(ICML  = {Int. Conf. Mach. Learn.})

@String(AAAI  = {AAAI})

@String(IJCAI = {IJCAI})

@String(CVPR  = {CVPR})

@String(ICCV  = {ICCV})

@String(ECCV  = {ECCV})

@String(NeurIPS = {NeurIPS})

@String(ICML  = {ICML})

@InProceedings{aicity2026track4,
  author    = {Tang, Zheng and Wang, Shuo and Anastasiu, David C. and Chang, Ming-Ching and others},
  title     = {The 10th {AI City Challenge}},
  booktitle = {ECCV Workshops},
  year      = {2026},
  address   = {Malm{"o}, Sweden}
}

@inproceedings{hu2025solution,
  author    = {Hu, Xiaoxing and Zheng, Tianlu and Yang, Kaicheng and Feng, Ziyong},
  title     = {The Solution to the {WWW25} Text-based Person Anomaly Search Challenge},
  booktitle = {Companion Proceedings of the ACM Web Conference 2025},
  year      = {2025},
  pages     = {1573--1575},
  doi       = {10.1145/3701716.3717651}
}

@inproceedings{nguyen2025hybrid,
  author    = {Nguyen, Tien-Huy and Tran, Huu-Loc and Phan-Nguyen, Huu-Phong and Dinh, Quang-Vinh},
  title     = {Hybrid, Unified and Iterative: A Novel Framework for Text-based Person Anomaly Retrieval},
  booktitle = {Companion Proceedings of the ACM Web Conference 2025},
  year      = {2025},
  pages     = {1576--1580},
  doi       = {10.1145/3701716.3717653}
}

@inproceedings{he2025efficient,
  author    = {He, Jiayi and Tang, Shengeng and Liu, Ao and Cheng, Lechao and Wu, Jingjing and Wei, Yanyan},
  title     = {Efficient Vision Language Model Fine-tuning for Text-based Person Anomaly Search},
  booktitle = {Companion Proceedings of the ACM Web Conference 2025},
  year      = {2025},
  pages     = {1568--1572},
  doi       = {10.1145/3701716.3717656}
}

@inproceedings{li2017person,
  title     = {Person Search with Natural Language Description},
  author    = {Li, Shuang and Xiao, Tong and Li, Hongsheng and Zhou, Bolei and Yue, Dayu and Wang, Xiaogang},
  booktitle = {Proceedings of the IEEE Conference on Computer Vision and Pattern Recognition (CVPR)},
  pages     = {1970--1979},
  year      = {2017}
}

@inproceedings{wang2020vitaa,
  title     = {ViTAA: Visual-Textual Attributes Alignment in Person Search by Natural Language},
  author    = {Wang, Zhe and Fang, Zhiyuan and Wang, Jun and Yang, Yezhou},
  booktitle = {Proceedings of the European Conference on Computer Vision (ECCV)},
  pages     = {402--420},
  year      = {2020}
}

@article{chen2022tipcb,
  title   = {TIPCB: A Simple but Effective Part-based Convolutional Baseline for Text-based Person Search},
  author  = {Chen, Yuhao and Zhang, Guoqing and Lu, Yujiang and Wang, Zhenxing and Zheng, Yuhui},
  journal = {Neurocomputing},
  volume  = {494},
  pages   = {171--181},
  year    = {2022}
}

@inproceedings{zhang2018deep,
  title     = {Deep Cross-Modal Projection Learning for Image-Text Matching},
  author    = {Zhang, Ying and Lu, Huchuan},
  booktitle = {Proceedings of the European Conference on Computer Vision (ECCV)},
  pages     = {686--701},
  year      = {2018}
}

@inproceedings{jiang2023irra,
  title     = {Cross-Modal Implicit Relation Reasoning and Aligning for Text-to-Image Person Retrieval},
  author    = {Jiang, Ding and Ye, Mang},
  booktitle = {Proceedings of the IEEE/CVF Conference on Computer Vision and Pattern Recognition (CVPR)},
  pages     = {2787--2797},
  year      = {2023}
}

@article{niu2020improving,
  title   = {Improving Description-based Person Re-identification by Multi-granularity Image-Text Alignments},
  author  = {Niu, Kai and Huang, Yan and Ouyang, Wanli and Wang, Liang},
  journal = {IEEE Transactions on Image Processing},
  volume  = {29},
  pages   = {5542--5556},
  year    = {2020}
}

@inproceedings{park2024plot,
  title     = {PLOT: Text-based Person Search with Part Slot Attention for Corresponding Part Discovery},
  author    = {Park, Jicheol and Kim, Dongwon and Jeong, Boseung and Kwak, Suha},
  booktitle = {Proceedings of the European Conference on Computer Vision (ECCV)},
  pages     = {474--490},
  year      = {2024}
}

@inproceedings{jing2020pose,
  title     = {Pose-guided Multi-granularity Attention Network for Text-based Person Search},
  author    = {Jing, Ya and Si, Chenyang and Wang, Junbo and Wang, Wei and Wang, Liang and Tan, Tieniu},
  booktitle = {Proceedings of the AAAI Conference on Artificial Intelligence (AAAI)},
  year      = {2020}
}

@inproceedings{bai2023rasa,
  title     = {RaSa: Relation and Sensitivity Aware Representation Learning for Text-based Person Search},
  author    = {Bai, Yang and Cao, Min and Gao, Daming and Cao, Ziqiang and Chen, Chen and Fan, Zhenfeng and Nie, Liqiang and Zhang, Min},
  booktitle = {Proceedings of the International Joint Conference on Artificial Intelligence (IJCAI)},
  pages     = {555--563},
  year      = {2023}
}

@inproceedings{radford2021learning,
  title     = {Learning Transferable Visual Models from Natural Language Supervision},
  author    = {Radford, Alec and Kim, Jong Wook and Hallacy, Chris and Ramesh, Aditya and Goh, Gabriel and Agarwal, Sandhini and Sastry, Girish and Askell, Amanda and Mishkin, Pamela and Clark, Jack and Krueger, Gretchen and Sutskever, Ilya},
  booktitle = {Proceedings of the International Conference on Machine Learning (ICML)},
  pages     = {8748--8763},
  year      = {2021}
}

@article{sun2023evaclip,
  title   = {EVA-CLIP: Improved Training Techniques for CLIP at Scale},
  author  = {Sun, Quan and Fang, Yuxin and Wu, Ledell and Wang, Xinlong and Cao, Yue},
  journal = {arXiv preprint arXiv:2303.15389},
  year    = {2023}
}

@misc{ilharco2021openclip,
  title        = {OpenCLIP},
  author       = {Ilharco, Gabriel and Wortsman, Mitchell and Wightman, Ross and Gordon, Cade and Carlini, Nicholas and Taori, Rohan and Dave, Achal and Shankar, Vaishaal and Namkoong, Hongseok and Miller, John and Hajishirzi, Hannaneh and Farhadi, Ali and Schmidt, Ludwig},
  year         = {2021},
  doi          = {10.5281/zenodo.7506443},
  note         = {Software release}
}

@inproceedings{zeng2022xvlm,
  title     = {Multi-Grained Vision Language Pre-Training: Aligning Texts with Visual Concepts},
  author    = {Zeng, Yan and Zhang, Xinsong and Li, Hang},
  booktitle = {Proceedings of the International Conference on Machine Learning (ICML)},
  pages     = {25994--26009},
  year      = {2022}
}

@article{llama3herd,
  title={The Llama 3 Herd of Models},
  author={Dubey, Abhimanyu and others},
  journal={arXiv preprint arXiv:2407.21783},
  year={2024}
}

@inproceedings{hu2022lora,
  title={LoRA: Low-Rank Adaptation of Large Language Models},
  author={Hu, Edward J. and Shen, Yelong and Wallis, Phillip and Allen-Zhu, Zeyuan and Li, Yuanzhi and Wang, Shean and Wang, Lu and Chen, Weizhu},
  booktitle={International Conference on Learning Representations},
  year={2022}
}

@article{qwen3technical,
  title={Qwen3 Technical Report},
  author={Qwen Team},
  journal={arXiv preprint arXiv:2505.09388},
  year={2025}
}

@article{qwen3vltechnical,
  title={Qwen3-VL Technical Report},
  author={Qwen Team},
  journal={arXiv preprint arXiv:2511.21631},
  year={2025}
}

@inproceedings{yang2024beyond,
  author    = {Yang, Shuyu and Wang, Yaxiong and Zhu, Li and Zheng, Zhedong},
  title     = {Beyond Walking: A Large-Scale Image-Text Benchmark for Text-Based Person Anomaly Search},
  booktitle = {Proceedings of the IEEE/CVF International Conference on Computer Vision (ICCV)},
  pages     = {11720--11730},
  year      = {2025}
}

@article{ju2025anomalylmm,
  title   = {AnomalyLMM: Bridging Generative Knowledge and Discriminative Retrieval for Text-based Person Anomaly Search},
  author  = {Ju, Hao and Zhang, Hu and Zheng, Zhedong},
  journal = {arXiv preprint arXiv:2509.04376},
  year    = {2025}
}

@inproceedings{sun2025filtering,
  title     = {From Data Deluge to Data Curation: A Filtering-WoRA Paradigm for Efficient Text-based Person Search},
  author    = {Sun, Jintao and Fei, Hao and Ding, Gangyi and Zheng, Zhedong},
  booktitle = {Proceedings of the ACM Web Conference (WWW)},
  pages     = {2341--2351},
  year      = {2025}
}

@inproceedings{cao2017realtime,
  title     = {Realtime Multi-Person 2D Pose Estimation Using Part Affinity Fields},
  author    = {Cao, Zhe and Simon, Tomas and Wei, Shih-En and Sheikh, Yaser},
  booktitle = {Proceedings of the IEEE Conference on Computer Vision and Pattern Recognition (CVPR)},
  pages     = {7291--7299},
  year      = {2017}
}

@inproceedings{sultani2018real,
  title     = {Real-World Anomaly Detection in Surveillance Videos},
  author    = {Sultani, Waqas and Chen, Chen and Shah, Mubarak},
  booktitle = {Proceedings of the IEEE Conference on Computer Vision and Pattern Recognition (CVPR)},
  pages     = {6479--6488},
  year      = {2018}
}

@misc{clip,
      title={Learning Transferable Visual Models From Natural Language Supervision}, 
      author={Alec Radford and Jong Wook Kim and Chris Hallacy and Aditya Ramesh and Gabriel Goh and Sandhini Agarwal and Girish Sastry and Amanda Askell and Pamela Mishkin and Jack Clark and Gretchen Krueger and Ilya Sutskever},
      year={2021},
      eprint={2103.00020},
      archivePrefix={arXiv},
      primaryClass={cs.CV},
      url={https://arxiv.org/abs/2103.00020}, 
}

@inproceedings{vismin,
  author    = {Awal, Rabiul and Ahmadi, Saba and Zhang, Le and Agrawal, Aishwarya},
  title     = {VisMin: Visual Minimal-Change Understanding},
  booktitle = {Advances in Neural Information Processing Systems (NeurIPS)},
  volume    = {37},
  year      = {2024},
  doi       = {10.52202/079017-3423}
}

@inproceedings{xie2026ssdc,
  author    = {Xie, Zequn and Luo, Guijin and Wang, Chuxin and Cai, Sihang and Jin, Tao and Zhao, Zhou and Tang, Yixuan},
  title     = {Bridging the Pose-Semantic Gap: A Cascade Framework for Text-Based Person Anomaly Search},
  booktitle = {Findings of the Association for Computational Linguistics: ACL 2026},
  pages     = {4040--4049},
  year      = {2026},
  publisher = {Association for Computational Linguistics}
}

@article{tran2026credence,
  title={CREDENCE: Claim Reduction for Decomposition \& Enhanced Credibility--Semantic Metrics and Convergence Analysis},
  author={Tran, Phuong Huu Vu and Mai, Thuan Duc and Le, Bach Xuan},
  journal={arXiv preprint arXiv:2606.19819},
  year={2026}
}

@inproceedings{verma2026selene,
    title = "{SELENE}: Selective and Evidence-Weighted {LLM} Debating for Efficient and Reliable Reasoning",
    author = "Verma, Akshay  and
      Gupta, Swapnil  and
      Gupta, Deepak  and
      Sircar, Prateek  and
      Pillai, Siddharth",
    editor = {Matusevych, Yevgen  and
      Eryi{\u{g}}it, G{\"u}l{\c{s}}en  and
      Aletras, Nikolaos},
    booktitle = "Proceedings of the 19th Conference of the {E}uropean Chapter of the {A}ssociation for {C}omputational {L}inguistics (Volume 5: Industry Track)",
    month = mar,
    year = "2026",
    address = "Rabat, Morocco",
    publisher = "Association for Computational Linguistics",
    url = "https://aclanthology.org/2026.eacl-industry.7/",
    doi = "10.18653/v1/2026.eacl-industry.7",
    pages = "95--104",
    ISBN = "979-8-89176-384-5"
}

@article{eo2025down,
  author  = {Eo, Sugyeong and Moon, Hyeonseok and Zi, Evelyn Hayoon and Park, Chanjun and Lim, Heuiseok},
  title   = {Debate Only When Necessary: Adaptive Multiagent Collaboration for Efficient {LLM} Reasoning},
  journal = {arXiv preprint arXiv:2504.05047},
  year    = {2025},
  url = {https://arxiv.org/abs/2504.05047}
}

@inproceedings{fan2026imad,
  author    = {Fan, Wei and Yoon, JinYi and Ji, Bo},
  title     = {{iMAD}: Intelligent Multi-Agent Debate for Efficient and Accurate {LLM} Inference},
  booktitle = {Proceedings of the AAAI Conference on Artificial Intelligence},
  volume    = {40},
  number    = {35},
  pages     = {29403--29411},
  year      = {2026},
  doi       = {10.1609/aaai.v40i35.40181}
}

@inproceedings{du2024improving,
  author    = {Du, Yilun and Li, Shuang and Torralba, Antonio and Tenenbaum, Joshua B. and Mordatch, Igor},
  title     = {Improving Factuality and Reasoning in Language Models through Multiagent Debate},
  booktitle = {Proceedings of the International Conference on Machine Learning (ICML)},
  year      = {2024}
}

@inproceedings{liang2024encouraging,
  author    = {Liang, Tian and He, Zhiwei and Jiao, Wenxiang and Wang, Xing and Wang, Yan and Wang, Rui and Yang, Yujiu and Shi, Shuming and Tu, Zhaopeng},
  title     = {Encouraging Divergent Thinking in Large Language Models through Multi-Agent Debate},
  booktitle = {Proceedings of the Conference on Empirical Methods in Natural Language Processing (EMNLP)},
  pages     = {17889--17904},
  year      = {2024}
}

@article{degroot1974consensus,
  author  = {DeGroot, Morris H.},
  title   = {Reaching a Consensus},
  journal = {Journal of the American Statistical Association},
  volume  = {69},
  number  = {345},
  pages   = {118--121},
  year    = {1974}
}

@article{genest1986combining,
  author  = {Genest, Christian and Zidek, James V.},
  title   = {Combining Probability Distributions: A Critique and an Annotated Bibliography},
  journal = {Statistical Science},
  volume  = {1},
  number  = {1},
  pages   = {114--135},
  year    = {1986}
}

@article{lin1991divergence,
  title={Divergence Measures Based on the Shannon Entropy},
  author={Lin, Jianhua},
  journal={IEEE Transactions on Information Theory},
  volume={37},
  number={1},
  pages={145--151},
  year={1991}
}

@book{cooke1991experts,
  title={Experts in Uncertainty: Opinion and Subjective Probability},
  author={Cooke, Roger M.},
  publisher={Oxford University Press},
  year={1991}
}

@article{qwen3vlembedding,
  title   = {Qwen3-VL-Embedding and Qwen3-VL-Reranker:
             A Unified Framework for State-of-the-Art
             Multimodal Retrieval and Ranking},
  author  = {Li, Mingxin and Zhang, Yanzhao and Long, Dingkun and
             Chen, Keqin and Song, Sibo and Bai, Shuai and
             Yang, Zhibo and Xie, Pengjun and Yang, An and
             Liu, Dayiheng and Zhou, Jingren and Lin, Junyang},
  journal = {arXiv preprint arXiv:2601.04720},
  year    = {2026}
}
\end{document}